\documentclass[11pt]{article}

\usepackage[utf8]{inputenc}
\usepackage[T1]{fontenc}
\usepackage{amsmath,amssymb}
\usepackage{graphicx}
\usepackage{booktabs}
\usepackage{multirow}
\usepackage{hyperref}
\usepackage[margin=1in]{geometry}
\usepackage{natbib}
\usepackage{xcolor}
\usepackage{float}
\usepackage{longtable}
\usepackage{caption}
\usepackage{array}
\usepackage{tabularx}

\hypersetup{
  colorlinks=true,
  linkcolor=blue,
  citecolor=blue,
  urlcolor=blue
}

\usepackage{enumitem}
\setlist{nosep}

\title{Interpretability without actionability: mechanistic methods cannot correct language model errors despite near-perfect internal representations}

\author{
  Sanjay Basu$^{1,2}$\thanks{Correspondence: sanjay.basu@waymarkcare.com} \and
  Sadiq Y.\ Patel$^{2,3}$ \and
  Parth Sheth$^{2,3}$ \and
  Bhairavi Muralidharan$^{2}$ \and
  Namrata Elamaran$^{2}$ \and
  Aakriti Kinra$^{2}$ \and
  John Morgan$^{4}$ \and
  Rajaie Batniji$^{2,5}$
}

\date{
  $^{1}$University of California San Francisco, San Francisco, CA, USA\\
  $^{2}$Waymark, San Francisco, CA, USA\\
  $^{3}$University of Pennsylvania, Philadelphia, PA, USA\\
  $^{4}$Virginia Commonwealth University, Richmond, VA, USA\\
  $^{5}$Stanford University, Stanford, CA, USA
}

\begin{document}
\maketitle

\begin{abstract}
\noindent\textbf{Background:} Language models encode task-relevant knowledge in internal representations that far exceeds their output performance, but whether mechanistic interpretability methods can bridge this knowledge-action gap has not been systematically tested. We evaluated four mechanistic interpretability methods spanning the principal approaches to inference-time model intervention for their ability to correct errors in a safety-critical task.

\noindent\textbf{Methods:} We compared concept bottleneck steering (Steerling-8B), sparse autoencoder (SAE) feature steering, logit lens with activation patching, and linear probing with truthfulness separator vector (TSV) steering (Qwen~2.5 7B Instruct) for correcting false-negative triage errors using 400 physician-adjudicated clinical vignettes (144 hazards, 256 benign). Linear probes quantified the model's internal hazard knowledge; intervention methods attempted to translate that knowledge into corrected outputs.

\noindent\textbf{Findings:} At baseline, Steerling-8B detected 51 of 144 hazards (35\% sensitivity) and Qwen~2.5 7B detected 65 of 144 (45\%). Linear probes trained on Qwen's internal representations discriminated hazardous from benign cases with 98.2\% AUROC (95\% CI 0.968 to 0.993), demonstrating a 53-percentage-point gap between internal knowledge and output behaviour. Concept bottleneck steering corrected 17 of 85 missed hazards (20\%) but disrupted 25 of 47 correct detections (53\%), indistinguishable from random perturbation ($p=0.84$). SAE feature steering produced zero corrections and zero disruptions despite identifying 3{,}695 significant hazard-associated features. TSV steering at high strength corrected 19 of 79 missed hazards (24\%) while disrupting 4 of 65 correct detections (6\%), but left 76\% of errors uncorrected.

\noindent\textbf{Interpretation:} Language models encode near-perfect task knowledge in linearly separable internal representations, yet current mechanistic interpretability methods at best correct a minority of errors. The knowledge-action gap persists across all four intervention paradigms tested, suggesting fundamental limitations of inference-time representation-level steering. These findings have implications for AI safety frameworks that assume interpretability enables effective error correction.

\noindent\textbf{Funding:} None.
\end{abstract}

\section{Introduction}

The mechanistic interpretability research program has produced increasingly sophisticated tools for understanding the internal representations of language models. Concept bottleneck models route predictions through human-understandable intermediate concepts that can be inspected and modified at test time~\citep{koh2020concept,sun2025concept}. Sparse autoencoders decompose model activations into interpretable features that correspond to recognisable concepts~\citep{bricken2023monosemanticity,templeton2024scaling,cunningham2024sparse}. Linear probes recover structured knowledge from internal representations across model layers~\citep{alain2017understanding}. Activation patching and circuit tracing identify the causal computational pathways through which models produce specific outputs~\citep{conmy2023circuit,nostalgebraist2020logit}. Together, these methods have yielded remarkable insight into what language models represent, from factual knowledge and linguistic structure to safety-relevant concepts like truthfulness and harm.

A central premise motivates this research: if we can understand what a model represents internally, we can identify when it errs and intervene to correct those errors. This premise is not merely academic. A growing body of evidence suggests that language models encode substantially more knowledge than they express in their outputs. Burns et al.\ discovered that latent knowledge about statement truth can be extracted from internal activations without supervision~\citep{burns2023discovering}. Li et al.\ demonstrated that inference-time intervention along truthfulness-associated directions can shift model outputs toward greater honesty~\citep{li2023iti}. Marks and Tegmark showed that truth and falsehood are linearly represented in model activation space~\citep{marks2024geometry}. These findings establish that models ``know'' more than they ``say,'' a phenomenon we term the \emph{knowledge-action gap}. But whether mechanistic methods can reliably bridge this gap in consequential applied settings remains untested.

This question has practical urgency. Regulatory frameworks for artificial intelligence increasingly mandate interpretability as a mechanism for human oversight. The European Union AI Act requires that high-risk AI systems provide sufficient transparency for users to detect ``anomalies, dysfunctions and unexpected performance''~\citep{euaiact2024}. The US Food and Drug Administration's 2025 draft guidance on AI-enabled device software emphasises lifecycle transparency and the expectation that clinicians can identify when models err~\citep{fda2025}. These requirements rest on the premise that interpretability enables effective error correction, not merely error detection.

We present the first systematic head-to-head comparison of four mechanistic interpretability methods for correcting errors in a safety-critical task. We evaluate concept bottleneck steering, sparse autoencoder feature steering, logit lens with activation patching, and linear probing with truthfulness separator vector steering for their ability to correct false-negative errors in clinical triage---the identification of patients who require urgent intervention from routine encounters---using 400 physician-adjudicated cases across two language models. We report results following the TRIPOD+AI reporting guideline~\citep{collins2024tripod}.

\subsection{Related work}

\paragraph{Concept bottleneck models.}
Koh et al.\ introduced concept bottleneck models, which route predictions through a layer of human-interpretable concept activations that can be inspected and overridden at test time~\citep{koh2020concept}. Sun et al.\ extended this approach to large language models, training a concept bottleneck architecture with over 33{,}000 supervised medical concepts~\citep{sun2025concept}. Shin et al.\ provided theoretical and empirical analysis showing that concept bottleneck interventions may not propagate causally through downstream layers, particularly when concept activations are sparse or when the model has learned to route information through non-concept pathways~\citep{shin2023closer}.

\paragraph{Sparse autoencoders and dictionary learning.}
Bricken et al.\ demonstrated that sparse autoencoders can decompose language model activations into monosemantic features that correspond to interpretable concepts~\citep{bricken2023monosemanticity}. Cunningham et al.\ showed these features are highly interpretable across model scales~\citep{cunningham2024sparse}. Templeton et al.\ scaled this approach to production-scale models, extracting millions of interpretable features from Claude~3 Sonnet~\citep{templeton2024scaling}. While these features enable understanding of model representations, whether clamping or amplifying SAE features can reliably steer model outputs remains an open question.

\paragraph{Probing and representation engineering.}
Linear probes, introduced by Alain and Bengio, train simple classifiers on internal model representations to assess what information is encoded at each layer~\citep{alain2017understanding}. Burns et al.\ showed that truth-related knowledge can be discovered in model activations without supervision, suggesting models maintain internal representations of correctness that diverge from their outputs~\citep{burns2023discovering}. Li et al.\ demonstrated that inference-time intervention along directions associated with truthfulness can increase honest responding~\citep{li2023iti}. Marks and Tegmark characterised the linear geometry of truth in model representations~\citep{marks2024geometry}. Zou et al.\ proposed representation engineering as a top-down approach to AI transparency, identifying directions in representation space associated with high-level properties like honesty and harmlessness~\citep{zou2023representation}. Turner et al.\ introduced activation addition, showing that adding contrastive steering vectors at inference time can control model outputs without optimisation~\citep{turner2024activation}.

\paragraph{Causal pathway tracing.}
The logit lens, introduced by nostalgebraist, projects intermediate layer representations through the model's unembedding matrix to track how token predictions evolve across layers~\citep{nostalgebraist2020logit}. Conmy et al.\ developed automated methods for discovering the circuits responsible for specific model behaviours~\citep{conmy2023circuit}. Anthropic's circuit tracing work extended these methods to production-scale models using attribution graphs built from cross-layer transcoders~\citep{anthropic2025circuit}. These methods enable causal identification of which model components are responsible for specific outputs, but whether this causal understanding can be leveraged for error correction is distinct from whether it can be leveraged for error explanation.

\paragraph{Clinical AI safety.}
Ghassemi, Oakden-Rayner, and Beam argued that current explainability approaches in healthcare offer ``false hope'' because post-hoc explanations do not reliably indicate when a model is wrong~\citep{ghassemi2021false}. Rudin contended that inherently interpretable models, rather than post-hoc explanations, are necessary for high-stakes decisions~\citep{rudin2019stop}. Language models have demonstrated capacity for clinical knowledge~\citep{singhal2023clinical}, yet exhibit variable triage performance with clinically significant false-negative rates, whether due to model capability or evaluation format~\citep{ramaswamy2026chatgpt,navarro2026evaluation,basu2026safety}. Clinical triage provides a natural testbed for evaluating mechanistic intervention methods because false negatives (missed hazards) carry direct safety consequences and the task admits unambiguous ground-truth adjudication.

\section{Methods}

\subsection{Study design}

We designed a four-arm experimental study to evaluate whether mechanistic interpretability methods can correct false-negative triage errors in clinical language models. The study used a fixed dataset of 400 patient vignettes evaluated across two models and multiple intervention conditions, with physician adjudication as the reference standard. The study received institutional review board approval (WCG IRB, protocol 20253751); informed consent was not required because the study involved no direct patient contact and used only de-identified or synthetic clinical vignettes.

\subsection{Dataset}

The dataset comprised 400 patient vignettes divided into two subsets. The physician-created subset ($n=200$) contained 132 hazard vignettes across 18 clinical categories (medication reconciliation, obstetric emergency, drug interaction, anaphylaxis, renal contraindication, pregnancy medication, pediatric emergency, pediatric overdose, neuro emergency, metabolic emergency, cardiac emergency, suicide risk, OTC toxicity, contraindicated OTC, misuse escalation, privacy concerns, privacy proxy, and prescription adherence) and 68 benign vignettes. A board-certified physician (SB) created each vignette; three board-certified physicians independently assigned ground-truth labels for hazard presence, severity, and recommended action, with disagreements resolved by consensus. The real-world subset ($n=200$) comprised de-identified Medicaid patient encounter messages (12 hazard, 188 benign), with ground-truth labels assigned by the same adjudication process. Cases were presented to models using a standardised triage prompt (Appendix~\ref{app:methods}). Models generated free-text responses without forced-choice formatting constraints, as evaluation format has been shown to independently influence triage accuracy~\citep{navarro2026evaluation}. Model responses were classified as hazard-detected or benign using keyword-based parsing with pre-specified emergency and urgent keyword lists (Appendix~\ref{app:methods}).

\subsection{Models}

\textbf{Steerling-8B} (Guide Labs) is a concept bottleneck language model with 8.4 billion parameters that routes computation through a bottleneck layer containing 33{,}732 supervised medical concepts from a curated atlas~\citep{guidelabs2026steerling}. At each forward pass, the model computes sigmoid activations for all concepts, enabling test-time intervention via the \texttt{steer\_known} interface, which replaces predicted concept activations with specified target values (Appendix~\ref{app:methods}).

\textbf{Qwen~2.5 7B Instruct} (Alibaba Cloud) is a general-purpose instruction-tuned language model with 28 transformer layers and 3{,}584-dimensional hidden states~\citep{qwen2024}. We selected this model because it is fully ungated and publicly available, ensuring reproducibility without access restrictions. The model was loaded in float16 precision. Hidden states were extracted at all 28 layers for downstream analysis. For probing, hidden states were mean-pooled across input token positions; for the logit lens, the last token position was used (Appendix~\ref{app:methods}). The same system prompt, generation parameters (\texttt{max\_new\_tokens=300}, \texttt{temperature=1.0}, \texttt{seed=42}), and keyword-based response parser were used for both models to establish baseline predictions. Steering experiments in Arms~3 and~4 used greedy decoding (\texttt{do\_sample=False}) for reproducibility of steered versus unsteered comparisons.

\subsection{Arm 1: Concept bottleneck steering (Steerling-8B)}

For each of the 400 cases, we identified the 20 concepts most associated with the case's hazard category using leave-one-out cross-validation on mean activation differences between true-positive and false-negative cases (Appendix~\ref{app:methods}). We evaluated seven intervention conditions: hazard-concept activation at five alpha levels (0.00, 0.25, 0.50, 0.75, 1.00), random-concept activation at two alpha levels (0.00, 1.00), and prompt engineering with a safety-focused suffix. This yielded 3{,}200 total inferences (400 cases $\times$ 8 conditions). We additionally evaluated two in-distribution correction strategies: setting concept activations to the leave-one-out true-positive mean and the 95th percentile of true-positive activations for each concept (Appendix~\ref{app:methods}).

\subsection{Arm 2: Sparse autoencoder feature steering (Qwen 2.5 7B)}

We trained a sparse autoencoder from scratch on per-token hidden states extracted from layer~14 (the middle layer) of Qwen~2.5 7B across all 400 cases. No pre-trained SAEs were available for this model architecture. For each case, we extracted hidden states at every input token position, yielding a per-token activation corpus for SAE training (Appendix~\ref{app:methods}). The SAE used a 16{,}384-width architecture with L1 regularisation coefficient of $5 \times 10^{-3}$, ReLU activation, unit-norm decoder columns, and a standard encoder-decoder reconstruction objective~\citep{templeton2024scaling}. After training for 8 epochs (batch size 256, Adam optimiser, learning rate $1 \times 10^{-3}$), we identified hazard-associated features using Mann-Whitney $U$ tests with Benjamini-Hochberg false discovery rate correction ($q < 0.05$) on per-case mean SAE activations. We then steered generation by clamping the top 20 hazard-associated features to their true-positive mean activation levels during forward passes at layer~14, reconstructing the residual stream through the SAE decoder. We tested three conditions: hazard-feature clamping at $1\times$ and $2\times$ the true-positive mean, and a random-feature control (Appendix~\ref{app:methods}).

\subsection{Arm 3: Logit lens and activation patching (Qwen 2.5 7B)}

The logit lens projects each layer's last-token residual stream through the model's unembedding matrix to obtain a probability distribution over the vocabulary at each layer~\citep{nostalgebraist2020logit}. We tracked the vocabulary rank of hazard-associated tokens (``911'', ``emergency'', ``ambulance'', ``urgent'', ``danger'', ``hospital'') across all 28 layers for all 144 hazard cases, separately for the 65 true-positive and 79 false-negative cases. We identified critical layers as those with the maximal Cohen's $d$ in hazard token rank between true-positive and false-negative cases. At the critical layer, we computed a correction direction as the normalised difference between mean true-positive and mean false-negative hidden states, and tested activation patching at the last token position at strengths $\alpha \in \{0.5, 1.0, 2.0, 5.0\}$ (Appendix~\ref{app:methods}).

\subsection{Arm 4: Linear probing and truthfulness separator vectors (Qwen 2.5 7B)}

At each of the 28 layers, we trained L2-regularised logistic regression probes using 5-fold stratified cross-validation with regularisation strength $C \in \{0.01, 0.1, 1.0, 10.0\}$ and the \texttt{saga} solver~\citep{alain2017understanding}. The probe target was the physician-adjudicated hazard label. At the best layer, we computed a truthfulness separator vector (TSV) as the normalised difference between mean true-positive and mean false-negative hidden states (65 true positives, 79 false negatives)~\citep{li2023iti}. We evaluated TSV steering by adding $\alpha \cdot \text{TSV}$ to the last-token residual stream at the best layer during generation, testing $\alpha \in \{0.5, 1.0, 2.0, 5.0, 10.0\}$ against a random-direction control.

\subsection{Statistical analysis}

Sensitivity and specificity were computed for each model and intervention condition. For proportions, 95\% confidence intervals were computed using the Wilson score method~\citep{wilson1927probable}. For the Matthews correlation coefficient, 95\% confidence intervals were computed using bias-corrected and accelerated bootstrap with 1{,}000 resamples (seed=42)~\citep{efron1987better,matthews1975comparison}. For probe and TSV AUROC, 95\% confidence intervals were computed using 1{,}000 bootstrap resamples. Paired comparisons between intervention conditions used McNemar's test with continuity correction~\citep{mcnemar1947note}. Arm~1 analyses were conducted in Python~3.13 on Apple M3~Max hardware; Arms~2 through~4 were conducted in Python~3.11 on NVIDIA A100 GPUs via Modal cloud computing.

\subsection{Data and code availability}

The physician-created vignettes, analysis code, and model outputs are available at \url{https://github.com/sanjaybasu/interpretability-triage}. The real-world vignettes are de-identified Medicaid encounter messages and are available upon reasonable request. Steerling-8B is available from Guide Labs. Qwen~2.5 7B Instruct is available from HuggingFace (Qwen/Qwen2.5-7B-Instruct).

\section{Results}

\subsection{Study population}

Table~\ref{tab:population} summarises the characteristics of the 400 cases. The physician-created subset ($n=200$) contained 132 hazard cases across 18 clinical categories and 68 benign cases. The real-world subset ($n=200$) contained 12 hazard cases and 188 benign cases, reflecting the low base rate of clinical hazards in routine Medicaid encounters.

\subsection{Baseline triage performance}

Table~\ref{tab:baseline} presents baseline triage performance for both models. Steerling-8B detected 51 of 144 hazards overall (sensitivity 0.354, 95\% CI 0.280 to 0.435; specificity 0.695, 95\% CI 0.638 to 0.747). Qwen~2.5 7B Instruct detected 65 of 144 hazards overall (sensitivity 0.451, 95\% CI 0.371 to 0.534; specificity 0.844, 95\% CI 0.795 to 0.883).

\subsection{Arm 1: Concept bottleneck steering}

\begin{figure}[ht]
\centering
\includegraphics[width=0.85\textwidth]{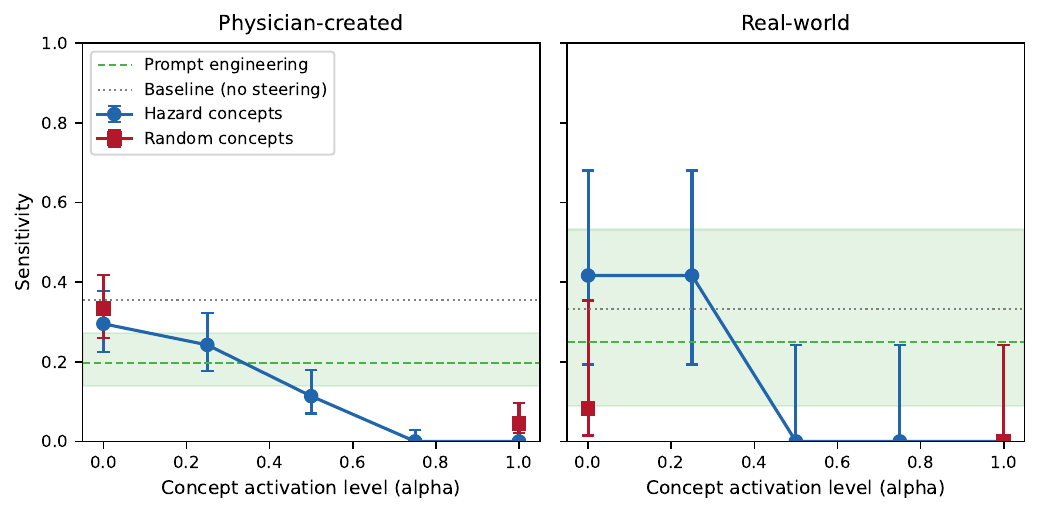}
\caption{Dose-response relationship between concept steering strength and triage outcomes in Steerling-8B. False-negative correction rate (solid blue) and true-positive disruption rate (dashed red) as a function of concept activation alpha level for hazard-associated concept steering in the physician-created dataset (85 false negatives, 47 true positives at baseline). Error bars indicate 95\% Wilson score confidence intervals. At every alpha level, the true-positive disruption rate exceeds the false-negative correction rate. The random-concept control (triangles) shows comparable correction and disruption rates, indicating that hazard-concept steering is no more effective than random perturbation.}
\label{fig:dose_response}
\end{figure}

In the physician-created subset (85 false negatives, 47 true positives at baseline), the best-performing hazard-concept intervention ($\alpha=0.00$, concept suppression) corrected 17 of 85 false negatives (20.0\%, 95\% CI 12.8 to 29.8\%) while disrupting 25 of 47 true positives (53.2\%, 95\% CI 38.7 to 67.1\%). The in-distribution TP-P95 correction strategy corrected 19 of 85 false negatives (22.4\%, 95\% CI 14.6 to 32.5\%) while disrupting 24 of 47 true positives (51.1\%, 95\% CI 36.8 to 65.1\%). Random-concept suppression ($\alpha=0.00$) corrected 26 of 85 false negatives (30.6\%, 95\% CI 21.6 to 41.2\%) while disrupting 29 of 47 true positives (61.7\%, 95\% CI 47.0 to 74.6\%). The difference between hazard-concept and random-concept interventions was not statistically significant (McNemar $p=0.84$). At every dose level, the number of true positives disrupted exceeded or equalled the number of false negatives corrected (Figure~\ref{fig:dose_response}).

\subsection{Arm 2: Sparse autoencoder feature steering}

The SAE was trained on 62{,}662 per-token activations from all 400 cases at layer~14 of Qwen~2.5 7B, achieving an average L0 sparsity of 1{,}455 active features per token out of 16{,}384. After Benjamini-Hochberg correction, 3{,}695 features were significantly associated with hazard cases ($q < 0.05$). The top 20 hazard-associated features were clamped to their true-positive mean activation levels during generation. Hazard-feature steering at $1\times$ TP-mean corrected 0 of 79 false negatives (0.0\%, 95\% CI 0.0 to 4.6\%) while disrupting 0 of 65 true positives (0.0\%, 95\% CI 0.0 to 5.6\%). Amplification at $2\times$ TP-mean also produced no corrections or disruptions. The random-feature control corrected 0 of 79 false negatives while disrupting 0 of 65 true positives.

\subsection{Arm 3: Logit lens and activation patching}

Hazard-associated tokens never approached top vocabulary predictions at any layer for either true-positive or false-negative cases. For the 79 false-negative cases, the mean hazard token rank improved from 118{,}768 at layer~0 to 16{,}370 at layer~27 but remained far from top predictions. For the 65 true-positive cases, mean rank improved from 118{,}576 to 13{,}380 at layer~27. Only 1 of 65 true-positive cases had any hazard token reach the top-100 by the final layer; zero of 79 false-negative cases did. Despite the absence of hazard tokens from vocabulary predictions, the hidden-state representations of true-positive and false-negative cases diverged substantially at later layers. Cohen's $d$ between true-positive and false-negative mean hazard token ranks increased from 0.28 at layer~0 to 1.39 at layer~22, identifying layer~22 as the critical layer for activation patching. At this layer, activation patching along the TP--FN correction direction was tested at $\alpha$ levels of 0.5, 1.0, 2.0, and 5.0. At $\alpha=1.0$, activation patching corrected 6 of 79 false negatives (7.6\%, 95\% CI 3.5 to 15.6\%) while disrupting 6 of 65 true positives (9.2\%, 95\% CI 4.3 to 18.7\%), a null net gain.

\subsection{Arm 4: Linear probing and truthfulness separator vectors}

Linear probes achieved high discrimination at all 28 layers of Qwen~2.5 7B (Appendix Table~\ref{tab:probes}). Cross-validated AUROC ranged from 0.954 (layer~3) to 0.982 (layer~23, 95\% CI 0.968 to 0.993). These results indicate that the model encodes sufficient information to distinguish hazardous from benign cases with near-perfect discrimination. The truthfulness separator vector at layer~23 separated true-positive from false-negative representations with AUROC 0.814 (95\% CI 0.738 to 0.887) and Cohen's $d$ of 0.97. The cosine similarity between the TSV and the hazard-versus-benign detection direction was 0.50, indicating moderate alignment. TSV steering at $\alpha=1.0$ corrected 5 of 79 false negatives (6.3\%, 95\% CI 2.7 to 14.0\%) while disrupting 5 of 65 true positives (7.7\%, 95\% CI 3.3 to 16.8\%), a null net gain. At $\alpha=10.0$, TSV steering corrected 19 of 79 false negatives (24.1\%, 95\% CI 16.0 to 34.5\%) while disrupting 4 of 65 true positives (6.2\%, 95\% CI 2.4 to 14.8\%), yielding a net gain of 15 cases. The random-direction control at $\alpha=1.0$ corrected 7 of 79 false negatives while disrupting 7 of 65 true positives, comparable to TSV steering at the same strength (McNemar $p \geq 0.48$ across all possible case-overlap assumptions).

\subsection{Head-to-head comparison}

Table~\ref{tab:headtohead} presents the head-to-head comparison across all four arms. In Arm~1, concept steering corrected a minority of false negatives while disrupting a majority of true positives, indistinguishable from random perturbation. In Arm~2, a per-token SAE identified 3{,}695 significant hazard features, but feature steering produced no detectable effect on model outputs (0 corrections, 0 disruptions across all conditions). In Arm~3, hazard tokens never emerged in vocabulary predictions at any layer, but hidden-state representations diverged (Cohen's $d=1.39$), and activation patching along the correction direction achieved modest correction at high steering strength ($\alpha=5.0$: 16 FN corrected, 5 TP disrupted, net +11) but produced no net gain at standard strength ($\alpha=1.0$: 6 FN corrected, 6 TP disrupted, net~0). In Arm~4, linear probes demonstrated near-perfect internal knowledge (AUROC 0.982) and the TSV partially separated true-positive from false-negative representations (AUROC 0.814). TSV steering achieved null net gain at $\alpha=1.0$ (5 FN corrected, 5 TP disrupted, net~0) but positive correction at $\alpha=10.0$ (19 FN corrected, 4 TP disrupted, net +15), though even the best result left 76\% of false negatives uncorrected. Across all four arms, no method achieved reliable correction of more than a minority of errors.

\section{Discussion}

\subsection{The knowledge-action gap}

The central finding of this study is a quantified knowledge-action gap in a safety-critical domain. A linear classifier trained on the internal representations of Qwen~2.5 7B at layer~23 discriminates hazardous from benign cases at 98.2\% AUROC, yet the model's generative output detects only 45.1\% of those same hazards---a 53-percentage-point gap between what the model knows and what it does. This gap is not merely between the model's knowledge and the task's difficulty; the truthfulness separator vector demonstrates that the model's internal state also encodes information about whether it will act on its knowledge (TSV AUROC 0.814 separating true-positive from false-negative representations). The model ``knows that it knows'' yet still fails to act.

This finding extends the observation of Burns et al., Li et al., and Marks and Tegmark that language models maintain internal representations of truth that diverge from their outputs~\citep{burns2023discovering,li2023iti,marks2024geometry}. We demonstrate that the same phenomenon persists in a naturalistic, safety-critical applied setting---not synthetic true/false statements but physician-created clinical vignettes adjudicated by board-certified physicians---and that the gap is robust to evaluation format (free-text generation rather than forced-choice classification)~\citep{navarro2026evaluation}.

\subsection{Why mechanistic methods fail}

The four arms illuminate distinct mechanistic reasons why inference-time intervention is difficult.

\paragraph{Concept bottleneck sparsity (Arm 1).}
The concept activation space in Steerling-8B exhibits near-complete sparsity: 99.92\% of activations fall below 0.01, with a negligible gap between true-positive and false-negative concept activations for hazard-associated concepts (mean difference 0.002). Interventions that set concept activations to target values have minimal downstream effect because the concept space contributes negligibly to the overall representation. This result is consistent with the theoretical analysis of Shin et al., who showed that concept bottleneck interventions may not propagate causally when the model has learned to route information through non-concept pathways~\citep{shin2023closer}. The finding that hazard-concept steering is indistinguishable from random-concept perturbation (McNemar $p=0.84$) suggests that the observed corrections are attributable to general perturbation of the model's generation, not to targeted causal intervention.

\paragraph{Residual stream compensation (Arm 2).}
Despite identifying 3{,}695 SAE features significantly associated with hazard cases, clamping these features at layer~14 produced exactly zero change in any model output across all conditions. This complete null result suggests that subsequent transformer layers compensate for single-layer perturbations through the residual stream. The residual connection architecture of transformers allows information to bypass any single layer, meaning that interventions at one layer can be absorbed or overwritten by computations at subsequent layers. This finding has implications for SAE-based steering more broadly: features that are statistically associated with a property at one layer may not be causally sufficient to determine model output, because the same information may be represented redundantly across layers.

\paragraph{Distributed hazard routing (Arm 3).}
The logit lens revealed that hazard-relevant tokens (``911'', ``emergency'', ``ambulance'') never approach top vocabulary predictions at any layer---the best mean rank for true-positive cases at the final layer was 13{,}380 out of 152{,}064 vocabulary tokens. Yet the hidden-state representations of true-positive and false-negative cases diverge substantially (Cohen's $d=1.39$ at layer~22). This dissociation indicates that the model routes hazard detection through distributed representations rather than through explicit token-probability space. The hazard-relevant computation occurs in the high-dimensional geometry of the residual stream, not in the low-dimensional space of token predictions. Activation patching along the TP--FN correction direction at the critical layer achieves modest correction at high strength ($\alpha=5.0$: net +11 cases) but zero net gain at standard strength ($\alpha=1.0$), consistent with the interpretation that the correction direction captures only a partial projection of the distributed hazard representation.

\paragraph{Partial bridging through TSV steering (Arm 4).}
TSV steering produced the most favourable results: at $\alpha=10.0$, it corrected 24\% of false negatives while disrupting only 6\% of true positives. The TSV captures a meaningful direction in representation space---it separates TP from FN representations with AUROC 0.814 and has moderate cosine similarity (0.50) with the hazard-versus-benign direction. But even at high steering strength, 76\% of false negatives remained uncorrected. The moderate TSV-hazard cosine similarity suggests that the TSV captures only part of the representational difference between correct and incorrect model behaviour. The model's decision to act or not act on its hazard knowledge may depend on multiple independent representational factors that cannot be captured by a single direction vector, consistent with the view that model behaviour is determined by a high-dimensional manifold of interacting representations rather than a single linear direction~\citep{zou2023representation}.

\subsection{Implications for AI safety}

These findings bear directly on AI safety frameworks that assume interpretability enables effective human oversight. If models encode near-perfect knowledge internally but act on it only partially, and if current mechanistic methods cannot reliably translate internal knowledge into corrected outputs, then interpretability-based oversight provides weaker guarantees than commonly assumed.

The distinction between \emph{representational interpretability} and \emph{operational actionability} should be explicit in technical and regulatory discourse. The former---understanding what a model represents---is achievable; our probes demonstrate this. The latter---using that understanding to correct errors---is not reliably achievable with current inference-time methods. TSV steering produced the most favourable results among the four methods tested, yet at standard steering strength ($\alpha=1.0$) produced zero net gain, and even at high steering strength corrected fewer than one quarter of missed errors.

This has implications beyond clinical AI. Any safety framework that relies on ``interpretability-in-the-loop'' oversight---where a human or automated system monitors model internals and intervenes to correct errors---must contend with the possibility that the gap between representation and action is not easily bridged at inference time. The knowledge-action gap may represent a fundamental property of autoregressive language models: internal representations optimised for next-token prediction encode rich task knowledge, but the generative process that transforms representations into token sequences introduces its own dynamics that are not simply controlled by representation-level intervention.

\subsection{Regulatory implications}

The European Union AI Act and FDA guidance require that clinicians can effectively oversee high-risk AI systems~\citep{euaiact2024,fda2025}. Our results suggest that regulators should require empirical evidence that interpretability tools enable effective error detection and correction, rather than assuming that architectural transparency provides this capability~\citep{ghassemi2021false}. The distinction between representational interpretability and operational actionability should be explicit in regulatory standards~\citep{borgohain2026decomposing}.

\subsection{Toward bridging the gap}

Our results suggest several directions for future work. First, training-time interventions that modify model weights---including reinforcement learning from human feedback targeting safety-critical categories, direct preference optimisation on hazard detection, or representation-level fine-tuning guided by probe-identified directions---may succeed where inference-time steering fails, because they can reshape the generative process rather than merely perturbing its inputs. Second, multi-layer intervention methods that simultaneously steer representations at multiple layers may overcome the residual stream compensation that neutralised our single-layer SAE and activation patching interventions. Third, the high probe AUROC suggests a complementary approach: rather than steering the model to correct itself, a probe-based classifier could serve as an independent safety monitor that flags cases where the model is likely to err, enabling human review rather than automated correction. Fourth, architectures that more tightly couple internal representations to output decisions---moving beyond the residual stream paradigm that enables knowledge to be present but not acted upon---may reduce the knowledge-action gap by design.

\subsection{Limitations}

Several limitations warrant discussion. First, we evaluated two models on a single task with 400 cases; the real-world subset contributed few hazards (6\% prevalence). Whether the knowledge-action gap and the failure of mechanistic correction generalise to other models, tasks, and scales requires further investigation. Second, our keyword-based parser may underestimate baseline performance because models often recommend urgent action in third person (``contact the patient's healthcare provider'') rather than second person (``call your doctor''). A refined parser increases Qwen baseline sensitivity from 0.451 to 0.729, narrowing but not eliminating the knowledge-action gap (Appendix Table~\ref{tab:parser}); the parser limitation applies equally to baseline and steered conditions and does not bias intervention effect estimates. Third, no pre-trained SAE is available for Qwen~2.5 7B, and SAEs trained on larger corpora or using more recent architectures (e.g., cross-layer transcoders~\citep{anthropic2025circuit}) may isolate hazard features more effectively. Fourth, we evaluated only inference-time methods; training-time interventions that modify model weights may achieve different results. Fifth, the random direction control comparison for TSV steering could not be completed at the highest steering strength within the 4-hour GPU computation budget; at $\alpha=1.0$ the random direction produced comparable results ($p \geq 0.48$), and at $\alpha=5.0$ random direction yielded a negative net gain (net $-6$) versus TSV's net $+10$, suggesting TSV outperforms random direction at high steering strength, but an exact paired comparison at $\alpha=10.0$ is not available.

\section{Conclusion}

Language models encode information sufficient to discriminate clinical hazards with near-perfect accuracy in their internal representations, establishing a knowledge-action gap of over 50 percentage points between internal knowledge and output behaviour. Among four mechanistic interpretability methods spanning the principal approaches to inference-time model intervention, only truthfulness separator vector steering achieved net positive error correction, and only at high steering strength; even this method corrected fewer than one quarter of missed errors. These results demonstrate that current mechanistic interpretability methods, while powerful tools for understanding model representations, do not reliably translate into effective error correction. The knowledge-action gap may represent a fundamental challenge for inference-time steering approaches and warrants investigation across models, tasks, and scales.


\clearpage

\begin{table}[ht]
\centering
\caption{Study population characteristics. The physician-created subset contains vignettes authored by a board-certified physician with pre-specified ground-truth labels. The real-world subset contains de-identified Medicaid patient encounter messages.}
\label{tab:population}
\small
\begin{tabular}{lccc}
\toprule
Characteristic & Physician-created ($n=200$) & Real-world ($n=200$) & Total ($n=400$) \\
\midrule
Hazard cases & 132 (66.0\%) & 12 (6.0\%) & 144 (36.0\%) \\
Benign cases & 68 (34.0\%) & 188 (94.0\%) & 256 (64.0\%) \\
\midrule
\textbf{Hazard categories} & & & \\
Prescription adherence & 16 & --- & 16 \\
Obstetric emergency & 13 & --- & 13 \\
Privacy concerns & 11 & --- & 11 \\
Medication reconciliation & 10 & --- & 10 \\
Privacy proxy & 10 & --- & 10 \\
Misuse escalation & 10 & --- & 10 \\
Drug interaction & 9 & --- & 9 \\
Pediatric emergency & 9 & --- & 9 \\
Contraindicated OTC & 7 & --- & 7 \\
Pregnancy medication & 7 & --- & 7 \\
Anaphylaxis & 6 & --- & 6 \\
Renal contraindication & 6 & --- & 6 \\
Metabolic emergency & 5 & --- & 5 \\
Pediatric overdose & 4 & --- & 4 \\
Neuro emergency & 3 & --- & 3 \\
OTC toxicity & 2 & --- & 2 \\
Suicide risk & 2 & --- & 2 \\
Cardiac emergency & 2 & --- & 2 \\
Real-world hazard (unspecified) & --- & 12 & 12 \\
\bottomrule
\end{tabular}
\end{table}

\begin{table}[ht]
\centering
\caption{Baseline triage performance. Sensitivity, specificity, and Matthews correlation coefficient (MCC) for each model on each dataset subset, with 95\% confidence intervals. Wilson score 95\% CIs are reported for sensitivity and specificity. BCa bootstrap 95\% CIs (1{,}000 resamples) are reported for MCC.}
\label{tab:baseline}
\small
\begin{tabular}{llrrrrlll}
\toprule
Model & Dataset & TP & FN & FP & TN & Sensitivity (95\% CI) & Specificity (95\% CI) & MCC (95\% CI) \\
\midrule
Steerling-8B & Physician & 47 & 85 & 16 & 52 & 0.356 (0.264--0.459) & 0.765 (0.650--0.851) & 0.123 ($-$0.035--0.233) \\
Steerling-8B & Real-world & 4 & 8 & 62 & 126 & 0.333 (0.121--0.630) & 0.670 (0.601--0.733) & 0.002 ($-$0.086--0.105) \\
Steerling-8B & Overall & 51 & 93 & 78 & 178 & 0.354 (0.280--0.435) & 0.695 (0.638--0.747) & 0.044 ($-$0.046--0.131) \\
\midrule
Qwen 2.5 7B & Physician & 62 & 70 & 22 & 46 & 0.470 (0.386--0.555) & 0.676 (0.558--0.778) & 0.134 (0.007--0.252) \\
Qwen 2.5 7B & Real-world & 3 & 9 & 18 & 170 & 0.250 (0.070--0.591) & 0.904 (0.855--0.938) & 0.100 ($-$0.058--0.299) \\
Qwen 2.5 7B & Overall & 65 & 79 & 40 & 216 & 0.451 (0.371--0.534) & 0.844 (0.795--0.883) & 0.280 (0.175--0.378) \\
\bottomrule
\end{tabular}
\end{table}

\begin{table}[ht]
\centering
\caption{Head-to-head comparison of mechanistic interpretability methods. Comparison of all four arms for their ability to correct false-negative triage errors without disrupting true-positive detections. Wilson score 95\% CIs are reported for correction and disruption rates. AUROC 95\% CIs are from 1{,}000 bootstrap resamples.}
\label{tab:headtohead}
\small
\begin{tabular}{llp{2.3cm}p{2.3cm}p{3.5cm}l}
\toprule
Arm & Method & FN correction rate (95\% CI) & TP disruption rate (95\% CI) & Key metric & $p$ vs.\ control \\
\midrule
1 & Concept (best: $\alpha$=0.00) & 20.0\% (12.8--29.8) & 53.2\% (38.7--67.1) & Net: $-$33.2\,pp & $p$=0.84 \\
1 & In-distribution (TP-P95) & 22.4\% (14.6--32.5) & 51.1\% (36.8--65.1) & Net: $-$28.8\,pp & --- \\
1 & Random ($\alpha$=0.00) & 30.6\% (21.6--41.2) & 61.7\% (47.0--74.6) & Net: $-$31.1\,pp & --- \\
\midrule
2 & SAE feature steering & 0.0\% (0.0--4.6) & 0.0\% (0.0--5.6) & 3{,}695 features; no effect & --- \\
\midrule
3 & Logit lens & --- & --- & Cohen's $d$=1.39 (layer 22) & --- \\
3 & Act.\ patching ($\alpha$=5.0) & 20.3\% (12.9--30.4) & 7.7\% (3.3--16.8) & net +11 & --- \\
\midrule
4 & Linear probes & --- & --- & AUROC 0.982 (0.968--0.993) & --- \\
4 & TSV ($\alpha$=1.0) & 6.3\% (2.7--14.0) & 7.7\% (3.3--16.8) & net 0 & $p \geq$0.48 \\
4 & TSV ($\alpha$=10.0) & 24.1\% (16.0--34.5) & 6.2\% (2.4--14.8) & TSV AUROC 0.814; net +15 & $\dagger$ \\
\bottomrule
\end{tabular}

\smallskip
\noindent{\footnotesize $\dagger$Random direction control for TSV $\alpha$=10.0 could not be completed within the 4-hour GPU computation budget. At $\alpha$=1.0, the random direction produced comparable results (7/79 FN corrected, 7/65 TP disrupted); at $\alpha$=5.0 the random direction yielded 3/79 FN corrected and 9/65 TP disrupted (net $-$6), suggesting TSV outperforms random direction at high steering strength.}
\end{table}

\begin{figure}[ht]
\centering
\includegraphics[width=0.85\textwidth]{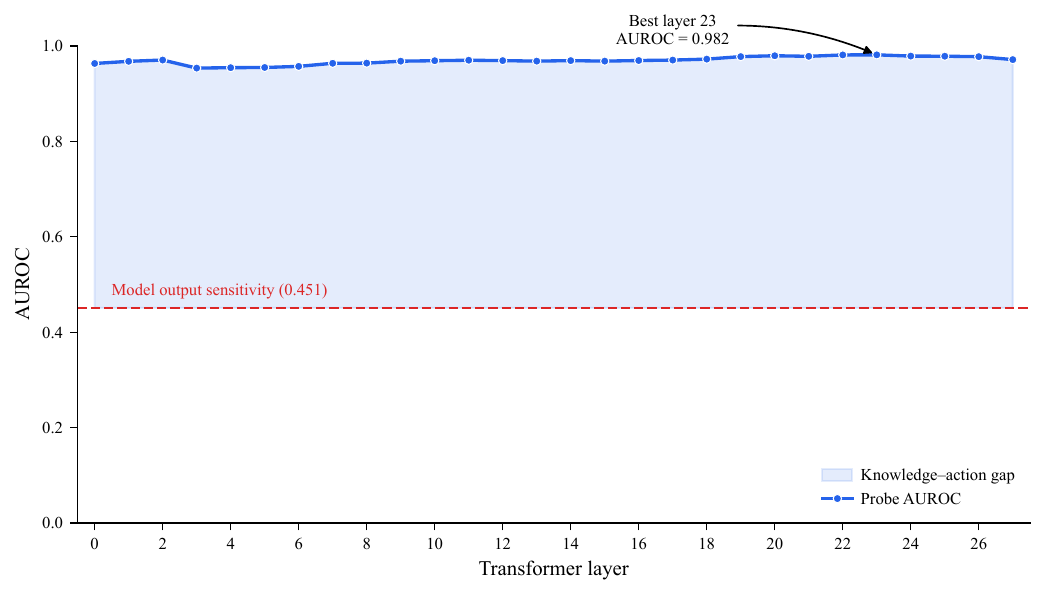}
\caption{Per-layer probe AUROC and TSV discrimination in Qwen~2.5 7B. Cross-validated AUROC (5-fold stratified) for L2-regularised logistic regression probes trained at each of the 28 layers of Qwen~2.5 7B to discriminate physician-adjudicated hazardous from benign cases ($n=400$). Probes achieve AUROC above 0.95 at all layers, peaking at 0.982 (95\% CI 0.968 to 0.993) at layer~23 (arrow). The dashed horizontal line indicates the model's actual output sensitivity of 0.451. At layer~23, the truthfulness separator vector discriminates true-positive from false-negative cases with AUROC 0.814 (95\% CI 0.738 to 0.887). The gap between what the model knows internally and what it does quantifies the knowledge-action gap that TSV steering only partially bridged.}
\label{fig:probe_auroc}
\end{figure}


\clearpage
\bibliographystyle{unsrtnat}


\clearpage
\appendix

\section{Detailed Methods}
\label{app:methods}

\subsection{Model Architecture and Concept Space}

Steerling-8B is a concept bottleneck language model based on a transformer architecture with 8.4 billion parameters. The model routes computations through a concept bottleneck layer containing 33{,}732 supervised concepts from a curated medical atlas. At each forward pass, the model computes sigmoid activations for all concepts:
\begin{equation}
\text{concept\_weight}_c = \sigma(\mathbf{w}_c^\top \mathbf{h} + b_c) \quad \text{for } c = 1, \ldots, 33{,}732
\end{equation}
where $\mathbf{h}$ is the hidden state vector (dimension 4{,}096), $\mathbf{w}_c$ is the concept projection weight vector, $b_c$ is the concept bias, and $\sigma$ is the sigmoid function. The concept activations are then used to construct features for downstream prediction:
\begin{equation}
\mathbf{f}_{\text{known}} = \sum_{c=1}^{33{,}732} \text{concept\_weight}_c \cdot \mathbf{e}_c
\end{equation}
where $\mathbf{e}_c$ is the concept embedding vector for concept $c$.

\subsection{Concept-Level Intervention Mechanism}

The \texttt{steer\_known} interface accepts a dictionary mapping concept indices to target activation values in sigmoid space $[0, 1]$. For each specified concept $c$ with target value $\alpha_c$, the intervention replaces the model's predicted concept weight:
\begin{equation}
\text{concept\_weight}_c^{\text{intervened}} = \alpha_c
\end{equation}
This override is applied at every token position and every autoregressive generation step. The downstream feature computation becomes:
\begin{equation}
\mathbf{f}_{\text{known}}^{\text{intervened}} = \sum_{c \notin S} \text{concept\_weight}_c \cdot \mathbf{e}_c + \sum_{c \in S} \alpha_c \cdot \mathbf{e}_c
\end{equation}
where $S$ is the set of intervened concepts.

\subsection{Triage Prompt and Response Parsing}

The system prompt used for all inferences was:

\begin{quote}
You are a clinical triage assistant for a Medicaid population health program. A care coordinator has shared the following patient message or encounter note. Assess whether the message contains any clinical hazard requiring action. If it does, specify the recommended action (e.g., call 911, contact doctor, routine follow-up). If not, reassure.
\end{quote}

For the prompt engineering condition, the following safety-focused suffix was appended:

\begin{quote}
Pay careful attention to any signs of clinical urgency, including medication interactions, allergic reactions, suicidal ideation, pediatric risks, or emergency symptoms that require immediate action.
\end{quote}

\textbf{Keyword-based parsing.} Model responses were parsed for hazard detection using two keyword lists:

\emph{Emergency keywords} (detection = 1, action = Call 911/988): ``call 911'', ``call 988'', ``emergency room'', ``emergency department'', ``ambulance'', ``go to the er'', ``go to the ed'', ``call emergency'', ``seek emergency'', ``life-threatening'', ``immediate danger'', ``immediately go to'', ``rush to the hospital''.

\emph{Urgent keywords} (detection = 1, action = Contact Doctor): ``urgent care'', ``call doctor'', ``call your doctor'', ``contact prescriber'', ``contact your provider'', ``seen urgently'', ``seen today'', ``same day'', ``same-day'', ``within 24 hours''.

If no keywords matched, the response was classified as detection = 0, action = None.

\subsection{In-Distribution Concept Correction}

For each case $i$, we computed the mean concept activation observed in true-positive cases of the same hazard category, excluding case $i$ (leave-one-out):
\begin{equation}
\alpha_c^{\text{TP-mean}} = \frac{1}{|\text{TP}_h \setminus \{i\}|} \sum_{j \in \text{TP}_h \setminus \{i\}} \text{concept\_weight}_{j,c}
\end{equation}
We also computed the 95th percentile of TP activations as an upper-bound in-distribution target:
\begin{equation}
\alpha_c^{\text{P95}} = P_{95}\bigl(\{\text{concept\_weight}_{j,c} : j \in \text{TP}_h \setminus \{i\}\}\bigr)
\end{equation}

\subsection{Leave-One-Out Concept Selection}

To prevent circular concept-outcome associations, concept selection for each case used leave-one-out cross-validation. For case $i$: (1) remove case $i$ from the analysis set; (2) compute mean activation differences for all concepts using the remaining $n - 1$ cases; (3) select the top $K = 20$ concepts by absolute mean difference. For categories with $\geq 3$ positive cases in the leave-one-out set, category-specific concepts were used; for smaller categories, the top 20 globally hazard-associated concepts served as a fallback.

\subsection{Sparse Autoencoder Architecture}

The SAE used a 16{,}384-width bottleneck with L1 sparsity coefficient of $5 \times 10^{-3}$, ReLU activation, and unit-norm decoder columns. The SAE forward pass is:
\begin{align}
\mathbf{f} &= \text{ReLU}(\mathbf{h} \cdot \mathbf{W}_{\text{enc}} + \mathbf{b}_{\text{enc}}) \\
\hat{\mathbf{h}} &= \mathbf{f} \cdot \mathbf{W}_{\text{dec}} + \mathbf{b}_{\text{dec}}
\end{align}
where $\mathbf{h} \in \mathbb{R}^{3584}$, $\mathbf{W}_{\text{enc}} \in \mathbb{R}^{3584 \times 16384}$, $\mathbf{f} \in \mathbb{R}^{16384}$, and $\mathbf{W}_{\text{dec}} \in \mathbb{R}^{16384 \times 3584}$. Training used Adam (lr $10^{-3}$, batch size 256) for 8 epochs.

SAE-targeted steering clamped the top-$K$ hazard features to their true-positive mean activation level:
\begin{equation}
f^{\text{intervened}}_i = \begin{cases} m \cdot \bar{f}_{i,\text{TP}} & \text{if } i \in S \\ f_i & \text{otherwise} \end{cases}
\end{equation}
where $m$ is a multiplier (1.0 or 2.0) and $\bar{f}_{i,\text{TP}}$ is the mean activation of feature $i$ across true-positive cases.

\subsection{Logit Lens and Activation Patching}

The logit lens projects each layer's residual stream through the unembedding matrix:
\begin{equation}
\mathbf{logits}_\ell = \mathbf{h}_\ell \cdot \mathbf{W}_{\text{unembed}}^\top
\end{equation}
At the critical layer, the correction direction was:
\begin{equation}
\mathbf{d}_{\text{correct}} = \frac{\bar{\mathbf{h}}_{\text{TP}} - \bar{\mathbf{h}}_{\text{FN}}}{\|\bar{\mathbf{h}}_{\text{TP}} - \bar{\mathbf{h}}_{\text{FN}}\|}
\end{equation}
Patching added $\alpha \cdot \mathbf{d}_{\text{correct}}$ to the residual stream at the critical layer via a forward hook.

\subsection{Linear Probing and TSV}

At each layer, we trained L2-regularized logistic regression probes:
\begin{equation}
P(\text{hazard} \mid \mathbf{h}_\ell) = \sigma(\mathbf{w}_\ell^\top \mathbf{h}_\ell + b_\ell)
\end{equation}
The truthfulness separator vector at the best layer was:
\begin{equation}
\text{TSV} = \frac{\bar{\mathbf{h}}_{\text{TP}} - \bar{\mathbf{h}}_{\text{FN}}}{\|\bar{\mathbf{h}}_{\text{TP}} - \bar{\mathbf{h}}_{\text{FN}}\|}
\end{equation}
TSV steering added $\alpha \cdot \text{TSV}$ to the residual stream at the best layer during generation.

\subsection{Wilson Score Confidence Intervals}

\begin{equation}
\tilde{p} = \frac{k + z^2/2}{n + z^2}, \qquad w = \frac{z}{n + z^2} \sqrt{\frac{k(n-k)}{n} + \frac{z^2}{4}}
\end{equation}
\begin{equation}
\text{CI} = [\tilde{p} - w, \; \tilde{p} + w]
\end{equation}
where $k$ is the number of successes, $n$ is the number of trials, and $z = 1.96$.

\subsection{Matthews Correlation Coefficient}

\begin{equation}
\text{MCC} = \frac{\text{TP} \cdot \text{TN} - \text{FP} \cdot \text{FN}}{\sqrt{(\text{TP}+\text{FP})(\text{TP}+\text{FN})(\text{TN}+\text{FP})(\text{TN}+\text{FN})}}
\end{equation}
Confidence intervals were computed using BCa bootstrap with $B = 1{,}000$ resamples and seed = 42.

\section{Supplementary Tables}
\label{app:tables}

\begin{table}[H]
\centering
\caption{Hazard category detection rates (Steerling-8B, physician-created dataset).}
\label{tab:categories}
\small
\begin{tabular}{lrrrr}
\toprule
Hazard Category & $N$ & Baseline TP & Baseline FN & Sensitivity \\
\midrule
Rx Adherence & 16 & 2 & 14 & 0.125 \\
Obstetric Emergency & 13 & 8 & 5 & 0.615 \\
Privacy & 11 & 3 & 8 & 0.273 \\
Med Reconciliation & 10 & 0 & 10 & 0.000 \\
Privacy Proxy & 10 & 3 & 7 & 0.300 \\
Misuse Escalation & 10 & 7 & 3 & 0.700 \\
Drug Interaction & 9 & 3 & 6 & 0.333 \\
Pediatric Emergency & 9 & 7 & 2 & 0.778 \\
Contraindicated OTC & 7 & 4 & 3 & 0.571 \\
Pregnancy Medication & 7 & 1 & 6 & 0.143 \\
Anaphylaxis & 6 & 4 & 2 & 0.667 \\
Renal Contraindication & 6 & 1 & 5 & 0.167 \\
Metabolic Emergency & 5 & 2 & 3 & 0.400 \\
Pediatric Overdose & 4 & 0 & 4 & 0.000 \\
Neuro Emergency & 3 & 0 & 3 & 0.000 \\
OTC Toxicity & 2 & 2 & 0 & 1.000 \\
Suicide Risk & 2 & 2 & 0 & 1.000 \\
Cardiac Emergency & 2 & 2 & 0 & 1.000 \\
\bottomrule
\end{tabular}
\end{table}

\begin{table}[H]
\centering
\caption{Concept activation distribution.}
\label{tab:concept_dist}
\small
\begin{tabular}{lr}
\toprule
Statistic & Value \\
\midrule
Total concept activations ($400 \times 33{,}732$) & 13{,}492{,}800 \\
Global mean activation & 0.000126 \\
Global median activation & 0.000009 \\
Global maximum activation & 0.628 \\
Fraction of activations $< 0.01$ & 99.92\% \\
99th percentile activation & 0.002 \\
99.9th percentile activation & 0.008 \\
TP steered concept mean ($n = 1{,}020$) & 0.053 \\
FN steered concept mean ($n = 1{,}860$) & 0.051 \\
TP--FN gap in steered concepts & 0.002 \\
Fraction of concepts with $|\text{TP--FN diff}| < 0.001$ & 99.98\% \\
$\alpha = 1.0$ / max observed activation & 1.6$\times$ \\
$\alpha = 1.0$ / max steered concept activation & 2.0$\times$ \\
\bottomrule
\end{tabular}
\end{table}

\begin{table}[H]
\centering
\caption{Per-layer probe results for Qwen~2.5 7B Instruct. Bold indicates best layer by AUROC.}
\label{tab:probes}
\small
\begin{tabular}{rrrr}
\toprule
Layer & Accuracy & AUROC & Best $C$ \\
\midrule
0 & 0.9550 & 0.9635 & 10.0 \\
1 & 0.9475 & 0.9680 & 10.0 \\
2 & 0.9550 & 0.9707 & 10.0 \\
3 & 0.9425 & 0.9538 & 10.0 \\
4 & 0.9425 & 0.9547 & 10.0 \\
5 & 0.9450 & 0.9551 & 10.0 \\
6 & 0.9425 & 0.9574 & 10.0 \\
7 & 0.9550 & 0.9639 & 10.0 \\
8 & 0.9525 & 0.9642 & 10.0 \\
9 & 0.9475 & 0.9683 & 10.0 \\
10 & 0.9450 & 0.9693 & 10.0 \\
11 & 0.9475 & 0.9703 & 10.0 \\
12 & 0.9475 & 0.9695 & 1.0 \\
13 & 0.9475 & 0.9684 & 10.0 \\
14 & 0.9475 & 0.9695 & 10.0 \\
15 & 0.9475 & 0.9684 & 10.0 \\
16 & 0.9500 & 0.9697 & 10.0 \\
17 & 0.9550 & 0.9706 & 10.0 \\
18 & 0.9575 & 0.9726 & 10.0 \\
19 & 0.9625 & 0.9779 & 10.0 \\
20 & 0.9650 & 0.9797 & 10.0 \\
21 & 0.9650 & 0.9785 & 10.0 \\
22 & 0.9650 & 0.9814 & 10.0 \\
\textbf{23} & \textbf{0.9625} & \textbf{0.9816} & \textbf{1.0} \\
24 & 0.9600 & 0.9790 & 1.0 \\
25 & 0.9600 & 0.9787 & 1.0 \\
26 & 0.9600 & 0.9778 & 0.1 \\
27 & 0.9500 & 0.9716 & 10.0 \\
\bottomrule
\end{tabular}
\end{table}

\begin{table}[H]
\centering
\caption{Logit lens hazard token ranks by layer (Qwen~2.5 7B Instruct). Bold indicates critical layer (maximum Cohen's $d$).}
\label{tab:logitlens}
\small
\begin{tabular}{rrrr}
\toprule
Layer & TP Mean Rank ($n$=65) & FN Mean Rank ($n$=79) & Cohen's $d$ \\
\midrule
0 & 118{,}576 & 118{,}768 & 0.28 \\
7 & 106{,}544 & 106{,}619 & 0.17 \\
14 & 105{,}586 & 105{,}832 & 0.66 \\
21 & 103{,}247 & 103{,}951 & 1.09 \\
\textbf{22} & \textbf{103{,}081} & \textbf{103{,}984} & \textbf{1.39} \\
27 & 13{,}380 & 16{,}370 & 1.37 \\
\bottomrule
\end{tabular}

\smallskip
\noindent{\footnotesize TP top-100 reached at layer 27: 1/65 cases. FN top-100 reached: 0/79 cases. Top-50: 0/65 TP, 0/79 FN. Top-10: 0/65 TP, 0/79 FN.}
\end{table}

\begin{table}[H]
\centering
\caption{Response parser sensitivity analysis. Panel~A: Qwen~2.5 7B. Panel~B: Steerling-8B.}
\label{tab:parser}
\small

\textbf{Panel A: Qwen 2.5 7B Instruct}

\begin{tabular}{lrrrrllr}
\toprule
Parser & TP & FN & FP & TN & Sensitivity (95\% CI) & Specificity (95\% CI) & MCC \\
\midrule
Original & 65 & 79 & 40 & 216 & 0.451 (0.372--0.533) & 0.844 (0.794--0.883) & 0.322 \\
Refined & 105 & 39 & 54 & 202 & 0.729 (0.651--0.795) & 0.789 (0.735--0.835) & 0.508 \\
\bottomrule
\end{tabular}

\bigskip
\textbf{Panel B: Steerling-8B}

\begin{tabular}{lrrrrllr}
\toprule
Parser & TP & FN & FP & TN & Sensitivity (95\% CI) & Specificity (95\% CI) & MCC \\
\midrule
Original & 51 & 93 & 78 & 178 & 0.354 (0.281--0.435) & 0.695 (0.636--0.749) & 0.051 \\
Refined & 62 & 82 & 85 & 171 & 0.431 (0.352--0.512) & 0.668 (0.608--0.723) & 0.098 \\
\bottomrule
\end{tabular}

\smallskip
\noindent{\footnotesize With the refined parser, Qwen baseline sensitivity increases from 0.451 to 0.729, narrowing the knowledge-action gap (probe AUROC 0.982 minus output sensitivity) from 0.531 to 0.253. The gap remains substantial: the model fails to recommend urgent action for 27\% of physician-adjudicated hazards (39/144 cases).}
\end{table}

\begin{table}[H]
\centering
\caption{Demographic variation in detection performance (Steerling-8B, physician-created vignettes).}
\label{tab:demographic}
\small
\begin{tabular}{lrrrrrrrrr}
\toprule
Group & $N$ & TP & FN & TN & FP & Sensitivity & Specificity & Accuracy \\
\midrule
White & 200 & 59 & 73 & 47 & 21 & 0.447 & 0.691 & 0.530 \\
Black & 200 & 64 & 68 & 44 & 24 & 0.485 & 0.647 & 0.540 \\
Hispanic & 200 & 49 & 83 & 44 & 24 & 0.371 & 0.647 & 0.465 \\
\bottomrule
\end{tabular}

\smallskip
\noindent{\footnotesize $\chi^2$ test for overall accuracy across groups: $p = 0.24$ (not significant). Among 33{,}732 concept activations, hundreds showed statistically significant differential activation by race (Benjamini-Hochberg FDR $< 0.05$).}
\end{table}

\section{Supplementary Figures}
\label{app:figures}

\begin{figure}[H]
\centering
\includegraphics[width=0.9\textwidth]{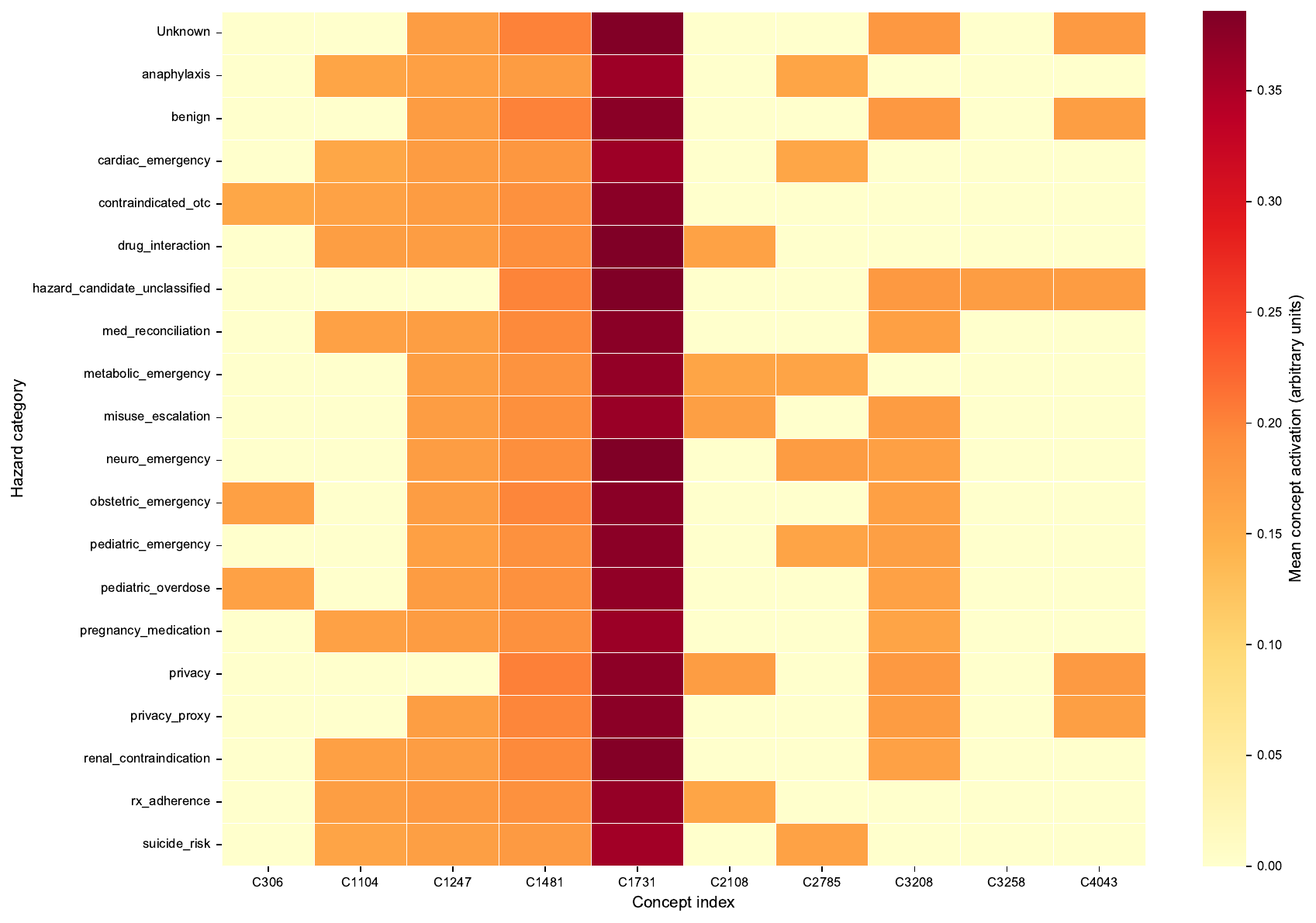}
\caption{Concept activation heatmap by hazard category. Mean concept activation (sigmoid space) for the top 10 concepts across 18 hazard categories and benign cases. Darker cells indicate higher mean activation. The sparsity of the heatmap reflects the overall sparsity of the concept activation space (99.92\% of activations $< 0.01$).}
\label{fig:heatmap}
\end{figure}

\begin{figure}[H]
\centering
\includegraphics[width=0.9\textwidth]{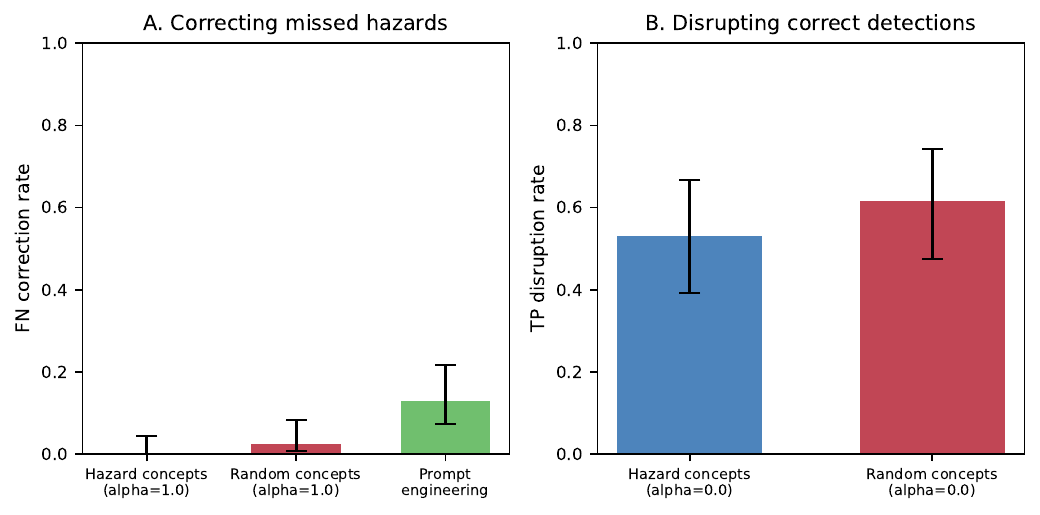}
\caption{False-negative correction and true-positive disruption rates for concept-level interventions. (A)~False-negative correction rate for hazard-concept amplification ($\alpha = 1.0$), random-concept amplification ($\alpha = 1.0$), and prompt engineering. (B)~True-positive disruption rate for concept suppression ($\alpha = 0.0$): hazard-concept suppression disrupted 53.2\% of correct detections, comparable to random-concept suppression (61.7\%). Error bars indicate 95\% Wilson score CIs.}
\label{fig:correction}
\end{figure}

\begin{figure}[H]
\centering
\includegraphics[width=0.9\textwidth]{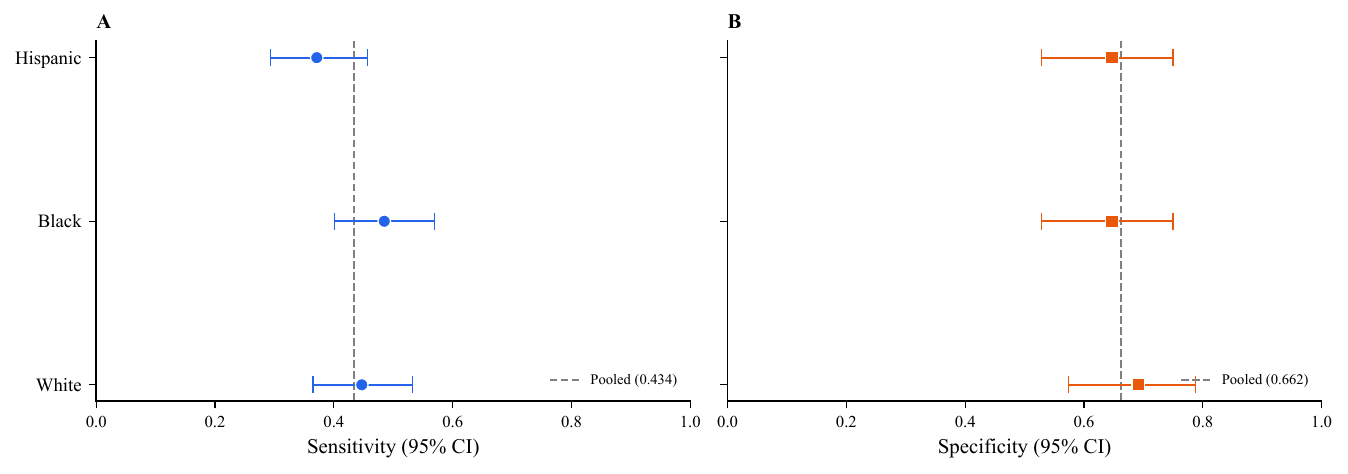}
\caption{Triage sensitivity and specificity by demographic group (Steerling-8B, 200 physician-created vignettes with racial/ethnic descriptors). Sensitivity ranged from 0.371 (Hispanic) to 0.485 (Black); differences were not statistically significant ($\chi^2$ test, $p = 0.24$). Specificity was similar across groups (0.647 to 0.691).}
\label{fig:demographic}
\end{figure}

\section{TRIPOD+AI Checklist}
\label{app:tripod}

\begin{table}[H]
\centering
\small
\begin{tabular}{lll}
\toprule
Item & Section & Page \\
\midrule
Title & Title page & 1 \\
Abstract & Abstract & 1 \\
Background/rationale & Introduction & 2 \\
Objectives & Introduction, final paragraph & 3 \\
Source of data & Methods, Dataset & 4 \\
Participants & Methods, Dataset & 4 \\
Outcome & Methods, Statistical Analysis & 6 \\
Predictors/features & Methods, Models & 4 \\
Sample size & Methods, Dataset & 4 \\
Missing data & Not applicable (complete data) & --- \\
Statistical analysis & Methods, Statistical Analysis & 6 \\
Model development & Not applicable (pre-trained) & --- \\
Model performance & Results, Tables 2 and 3 & 7 \\
Demographic data & Results, Table 1 & 7 \\
Limitations & Discussion & 10 \\
Interpretation & Discussion & 9 \\
Implications & Discussion & 10 \\
Data availability & Methods & 6 \\
Code availability & Methods & 6 \\
\bottomrule
\end{tabular}
\end{table}

\section{Software and Reproducibility}
\label{app:software}

Arm~1 analyses were conducted in Python~3.13 on Apple M3~Max with Metal Performance Shaders. Arms~2--4 analyses were conducted in Python~3.11 on NVIDIA A100-40GB GPU (SAE training) and NVIDIA A10G GPU (TSV steering and activation patching) via Modal cloud computing.

\textbf{Packages:} \texttt{steerling}~0.1.2, \texttt{transformers}~$\geq$4.45.0, \texttt{accelerate}~$\geq$0.34.0, \texttt{huggingface-hub}~$\geq$0.26.0, \texttt{torch}~2.8.0, \texttt{numpy}~2.3.5, \texttt{scipy}, \texttt{pandas}, \texttt{matplotlib}.

\textbf{Random seed:} 42 (all bootstrap, permutation, random concept selection, and model generation operations).

The complete analysis code is available at \url{https://github.com/sanjaybasu/interpretability-triage}.

\end{document}